\documentclass[letterpaper, 10 pt, journal, twoside]{ieeetran}

\usepackage{graphicx} %for figures
\graphicspath{{Figures/}}
\usepackage{url} 
\usepackage{amsmath}
\usepackage{amssymb}
\usepackage{gensymb}
\usepackage{lscape}
\usepackage{caption}
\usepackage{subcaption}
\usepackage{cite} %handles multiple citations
\usepackage{booktabs} %for professional tables
\usepackage[hidelinks]{hyperref} %for hyperlinks
\usepackage{xcolor,colortbl}
\usepackage[english]{babel}
\usepackage{csquotes}
\usepackage{notoccite}
\usepackage{blindtext}
\usepackage{listings}
\usepackage{xcolor}
\usepackage{multirow}
\usepackage{tabulary}
\usepackage[printonlyused]{acronym}
\usepackage{pdflscape}
\usepackage{latexsym}
\usepackage[bottom]{footmisc}
\usepackage{textcomp}
\usepackage{epstopdf}
\usepackage{soul}
\usepackage[ruled,vlined]{algorithm2e}
\usepackage{bm}

\begin{document}
%
% paper title
% Titles are generally capitalized except for words such as a, an, and, as,
% at, but, by, for, in, nor, of, on, or, the, to and up, which are usually
% not capitalized unless they are the first or last word of the title.
% Linebreaks \\ can be used within to get better formatting as desired.
% Do not put math or special symbols in the title.
\title{Feedback Control of Millimeter Scale Pivot Walkers Using Magnetic Actuation}

% author names and affiliations
% transmag papers use the long conference author name format.

\author{Ehab Al Khatib, Pouria Razzaghi, and Yildirim Hurmuzlu$^*$

\thanks{Ehab Al Khatib, Pouria Razzaghi, and Yildirim Hurmuzlu are from the Department of Mechanical Engineering, Southern
	Methodist University, Dallas, TX 75275, U.S.A. (corresponding author:
	hurmuzlu@lyle.smu.edu}}

% The paper headers
%\markboth{IEEE ROBOTICS AND AUTOMATION LETTERS}%
{}
% The only time the second header will appear is for the odd numbered pages
% after the title page when using the twoside option.
% 
% *** Note that you probably will NOT want to include the author's ***
% *** name in the headers of peer review papers.                   ***
% You can use \ifCLASSOPTIONpeerreview for conditional compilation here if
% you desire.

% If you want to put a publisher's ID mark on the page you can do it like
% this:
%\IEEEpubid{0000--0000/00\$00.00~\copyright~2015 IEEE}
% Remember, if you use this you must call \IEEEpubidadjcol in the second
% column for its text to clear the IEEEpubid mark.

% use for special paper notices
%\IEEEspecialpapernotice{(Invited Paper)}

% for Transactions on Magnetics papers, we must declare the abstract and
% index terms PRIOR to the title within the \IEEEtitleabstractindextext
% IEEEtran command as these need to go into the title area created by
% \maketitle.
% As a general rule, do not put math, special symbols or citations
% in the abstract or keywords.
\IEEEtitleabstractindextext{%
\begin{abstract}
An external magnetic field can be used to remotely control small-scaled robots, making them promising candidates for diverse biomedical and engineering applications. In previous study, we showed that our magnetically actuated millirobot is highly agile and can perform a variety of locomotive tasks such as pivot walking, tumbling, and tapping in a horizontal plane. In this study, we focus on the controlling of the locomotion outcomes of this millirobot in the pivot walking mode. A mathematical model of the system is developed and the kinematic model is derived. The role of the sweep and tilt angles in robot's motion is also investigated.

We propose two controllers to regulate the gait of the pivot walker. The first one is a proportional-geometric controller, which determines the correct pivot point that the millirobot should use. Then, it regulates the angular velocity proportionally based on the error between the center of the millirobot and the reference trajectory. The second controller is based on a gradient descent optimization technique, which expresses the control action as an optimization problem. These control algorithms enable the millirobot to generate stable gait while tracking a desired trajectory. A low-cost high-performance magnetic actuator is built to validate the proposed controllers. We conduct a set of different experiments and simulation runs to establish the effectiveness of proposed controllers for different sweep and tilt angles in terms of the tracking error. The two controllers exhibit an appropriate performance, but it is observed that gradient descent based controller yields faster convergence time, smaller tracking error, and fewer number of steps. Finally, we perform an extensive experimentally parametric analysis on the effect of the sweep angle, tilt angle, and step time on the tracking error. As we expect, the optimization-based controller outperforms the geometric based controller.

\end{abstract}

% Note that keywords are not normally used for peerreview papers.
}

% make the title area
\maketitle

% To allow for easy dual compilation without having to reenter the
% abstract/keywords data, the \IEEEtitleabstractindextext text will
% not be used in maketitle, but will appear (i.e., to be "transported")
% here as \IEEEdisplaynontitleabstractindextext when the compsoc 
% or transmag modes are not selected <OR> if conference mode is selected 
% - because all conference papers position the abstract like regular
% papers do.
\IEEEdisplaynontitleabstractindextext
% \IEEEdisplaynontitleabstractindextext has no effect when using
% compsoc or transmag under a non-conference mode.

% For peer review papers, you can put extra information on the cover
% page as needed:
% \ifCLASSOPTIONpeerreview
% \begin{center} \bfseries EDICS Category: 3-BBND \end{center}
% \fi
%
% For peerreview papers, this IEEEtran command inserts a page break and
% creates the second title. It will be ignored for other modes.
\IEEEpeerreviewmaketitle

\section{Introduction}\IEEEPARstart{B}{iologically} inspired locomotion and the growth in need for mobility in challenging and unknown environments have motivated many researchers to design different miniature robots. These robots can be listed as conventional wheeled, crawling and snake-like, legged, and hopper. Recently,  traversability in irregular environments has become an important feature in the design of robots. Small-scale robots show great potential capabilities in maneuverability and reachability in such environments. Also, these robots can benefit from untethered actuation mechanisms and free from limitations imposed by onboard actuators and power sources. Untethered tiny robots can be actuated by external forces, such as electromagnetic fields \cite{palagi2018bioinspired,shahrokhi2019exploiting}, acoustic waves \cite{aghakhani2020acoustically}, and light \cite{shahrokhi2016object,sridhar2018light,shahsavan2020bioinspired}. In this paper, we present a millimeter-scale magnetic robot that can be steered by an external magnetic field. 

Small scaled robots actuated and controlled by external magnetic fields can be used in different applications such as accessing and manipulating objects in tight spaces. Various locomotion modes has been developed for these types of robots such as \textit{Pulling/Pushing} \cite{abbott2009should}, \textit{Tumbling/Rolling} \cite{tung2013noncontact}, \textit{Helical Thrusting} \cite{bozuyuk2018light}, \textit{Swimming} \cite{diller2014continuously}, \textit{Crawling} \cite{hu2018small}, \textit{Stick-Slipping} \cite{pawashe2009modeling}, and \textit{Pivot Walking} \cite{al2020magnetically}. These locomotions can be carried out by rigid or soft mechanisms \cite{de2019review}. 

Here, we summarize these different motion modes. Steering a permanent magnet can generate a magnetic force and subsequently pull or push a magnetic body on a dry surface or in a fluid environment. Diller \textit{et al.} \cite{diller2013independent} controlled a sub-millimeter robots by using a magnetic gradient pulling method. The resulting tumbling motion exhibited a more controllable mode without significant slippage. Bi \textit{et al.} \cite{bi2019tumbling} presented a microrobot that is controlled by an external magnetic field. A permanent magnet was rotated beneath the workspace and regulated a microrobot to tumble and travel through a narrow channel. Alternatively, when a rotating magnetic field is applied to a helical robot submerged in a fluid, it will move in the direction of the helical axis. Bozuyuk \textit{et al.} \cite{bozuyuk2018light} 3D printed a double-helical microswimmer made from a magnetic polymer composite. The motion of this robot was controlled using light in order to perform drug payload carry and release tasks.

In addition, robots can swim in a liquid medium by producing a wavy motion of their bodies. Zhang and Diller \cite{zhang2018untethered} designed a flexible magnetic sheet, which can swim in a liquid. Using the same technique, one can also generate crawling motion on dry surfaces. Accordingly, Hu \textit{et al.} \cite{hu2018small} developed a non-uniform magnetized soft millimeter-scale robot and controlled it to move on a solid surface, swim inside and on the surface of a liquid, and crawl in a channel.

Kashki \textit{et al.} \cite{kashki2016pivot} introduced an inertially actuated robot that can produce a type of wobbling gate, which they called ``pivot walking''. This type of locomotion is based on using two spinners that cause the robot to successively pivot and spin about two distinct points. It turns out, one can generate similar motions using magnetic instead of inertial actuation. Using magnetic actuation makes it possible to significantly reduce the size of pivot walkers because the burden of having on-board actuators is now moved elsewhere. This type of magnetically actuated locomotion was used by Dong and Sitti \cite{7989782} to regulate the motion of a microgripper. They controlled the gripper to reach and grip an object, then move it and release it at the desired destination. 

In our recent study \cite{al2020magnetically}, we proposed a millirobot, with two permanent magnets mounted at each end of a rectangular shaped body. This millirobot can perform several locomotion modes, including pivot walking, tumbling, and rolling. The control object was an open loop control mode, where sequential clockwise and counter-clockwise magnetic torque were applied to the millirobot to generate locomotion. We discovered that the most controllable motion among these modes was pivot walking. Closed-loop position control of this millirobot is a challenging task. In this study, we investigate the kinematic model of the pivot walking mode. We propose two control algorithms that track time-dependent desired trajectories based on millirobot's kinematics. We verify our proposed controllers through simulation and experimental results. 

The first scheme is a proportional-geometric controller, which moves the center of the millirobot to the desired location at that time instant. The controller performs this task by taking successive steps alternating between the two pivot points. During each step, based on the distances between the pivot points and the desired position, one of the pivot points is chosen. Then, the millirobot swings about the chosen pivot with a velocity that is proportional to the distance between the center of the millirobot and the desired point.

The second control scheme is based on a gradient descent optimization technique, where the control problem is expressed as an optimization problem. The control inputs are the parameters that need to be optimized to reduce the error between the center of the millirobot and the desired trajectory. These control algorithms enable the millirobot to walk and track an arbitrary trajectory.

The locomotion of the millirobot depends on three parameters, sweep angle, tilt angle, and generating path step time. We conduct extensive experiments to perform parametric analysis on the effects of varying these parameters on the trajectory tracking error.   

This paper is organized as follows. In Section \ref{section2}, a kinematic model of the pivot walker millirobot is derived. Section \ref{section3} presents the locomotion analysis. Two control algorithms are presented in Section \ref{section4}. We carry out numerical simulation runs to validate the performance of the proposed controllers in Section \ref{section5}. The experimental results and parametric analysis on pivot walking mode are presented in Section \ref{section6}, and the conclusion is in Section \ref{section7}.

\section{Kinematic model} \label{section2}

\begin{figure}[!b]
	\centering
	\includegraphics[width=1\columnwidth]{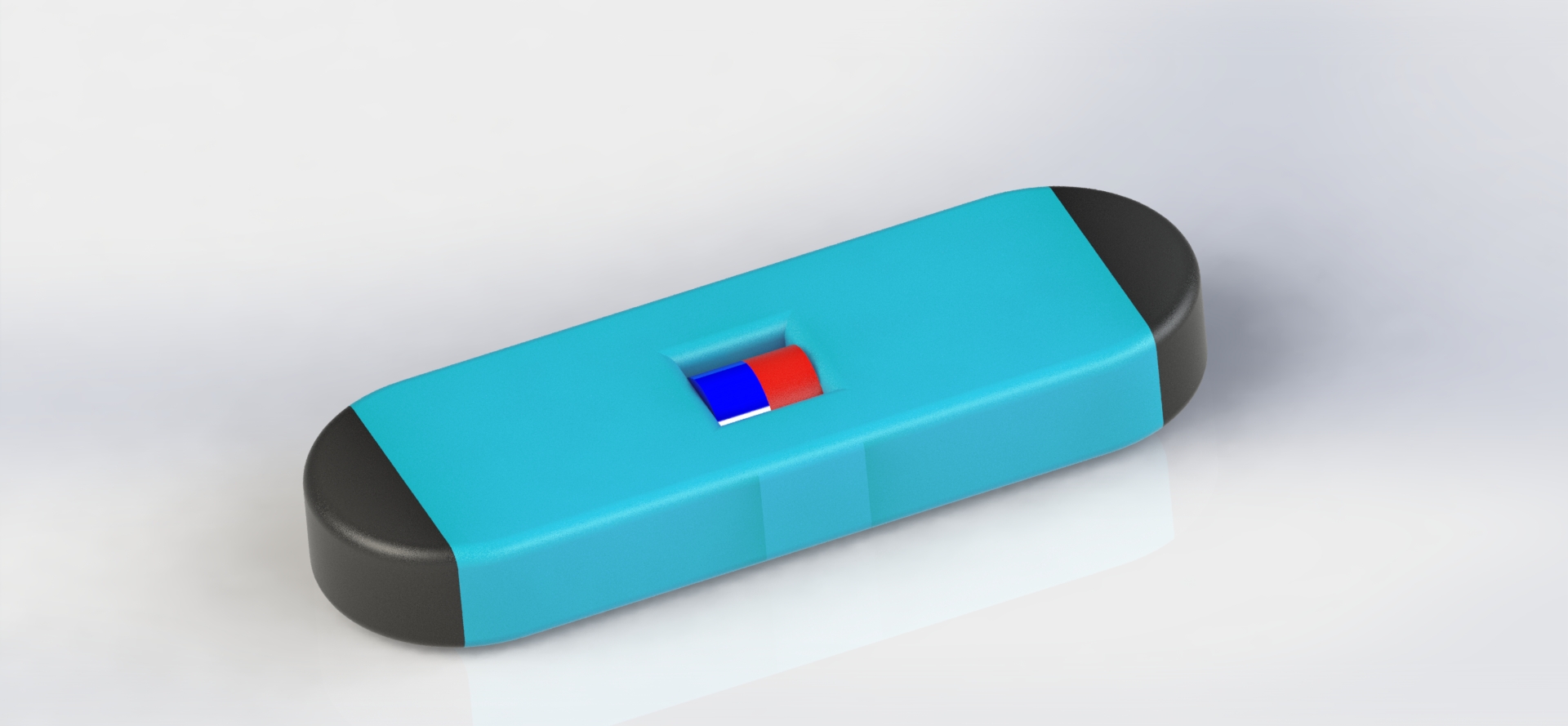}
	\caption{\small{CAD design of a pivot walking millirobot. A permanent magnet is embedded at the center. The red and blue sides are represented the north and south magnetic poles respectively. }}
	\label{pivotWalker15}
\end{figure}
\normalsize

\begin{figure}[!t]
	\centering
	\includegraphics[width=0.75\columnwidth]{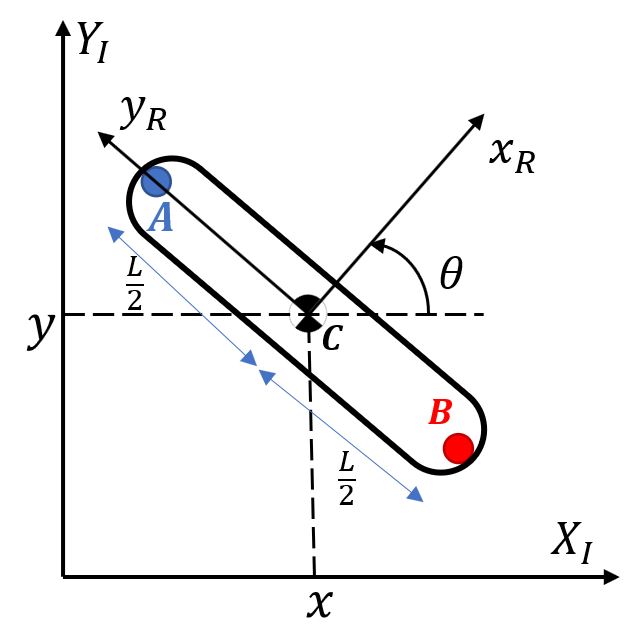}
	\caption{\small{The global reference frame and the millirobot local reference frame. }}
	\label{referenceFrame}
\end{figure}

A millirobot is composed of a rigid link with a permanent magnet embedded in the center as shown in Fig.~\ref{pivotWalker15}. This millirobot is capable of forming a pivot point at two ends by manipulating the surrounding magnetic field \cite{al2020magnetically}. The magnetic millirobot is actuated by an external magnetic field $\bar{B}$. The induced magnetic torque will rotate and align the millirobot's permanent magnetic $\bar{M}$ with the external magnetic field. The applied torque on the millirobot can be expressed as follows:
\begin{equation}
\bar{\tau}=\bar{M} \times \bar{B}
\end{equation}
where $\bar{\tau}$ is the induced magnetic torque vector. As shown in Fig.~\ref{referenceFrame}, the pivot points are labeled as \textbf{\textit{A}} and \textbf{\textit{B}}. A pivot point is formed when one end is pressed down while the other end is lifted up by a tilt angle $(\alpha)$ by applying the induced magnetic field in that direction. 

\normalsize

Derivation of the kinematics of the pivot walker is based on the assumption that the millirobot is constructed as a rigid body. To identify the millirobot position on the plane, two reference frames are used; the global and the local reference frames as shown in Fig.~\ref{referenceFrame}. The axes $X_I$ and $Y_I$ define an arbitrary inertial basis in the global reference frame. To specify the position of the millirobot, a point \textbf{\textit{C}} on the center of the millirobot is chosen. The basis $\{x_C,y_C\}$ defines the millirobot's local reference frame that passes through \textbf{\textit{C}}. The position of \textbf{\textit{C}} in the global reference frame is specified by coordinates $x$ and $y$, and the angle between the two frames is given by $\theta$. The robot position is then fully identified by the three variables $x,y$, and $\theta$. A 3$\times$1-vector $\bar{q}_I$ is defined to describe the millirobot state in the global frame and $\bar{q}_R$ in the local frame.
\begin{equation}
\bar{q}_I=\begin{bmatrix}
x & y & \theta 
\end{bmatrix}^T
\label{eq:KM1}
\end{equation}
The transformation between the millirobot frame and inertial frame is described as follows:
\begin{equation}
\dot{\bar{q}}_R = R(\theta )\dot{\bar{q}}_I
\label{eq:KM3}
\end{equation}
\begin{equation}
\textbf{R}(\theta )=\begin{bmatrix}
\cos \theta & \sin \theta  & 0\\ 
- \sin \theta & \cos \theta  &0 \\ 
0& 0 &1 
\end{bmatrix}
\label{eq:KM2}
\end{equation}
where, $\textbf{R}(\theta)$ is the rotation matrix. Based on Eq.~(\ref{eq:KM3}) the kinematics of the millirobot in the inertial frame can be written as follows:
\begin{equation}
\dot {\bar{q}}_I = {\textbf{R}(\theta)}^{-1} \dot{\bar{q}}_R
\label{eq:KM7}
\end{equation}

\begin{figure}[!t]
	\centering
	\includegraphics[width=0.9\columnwidth]{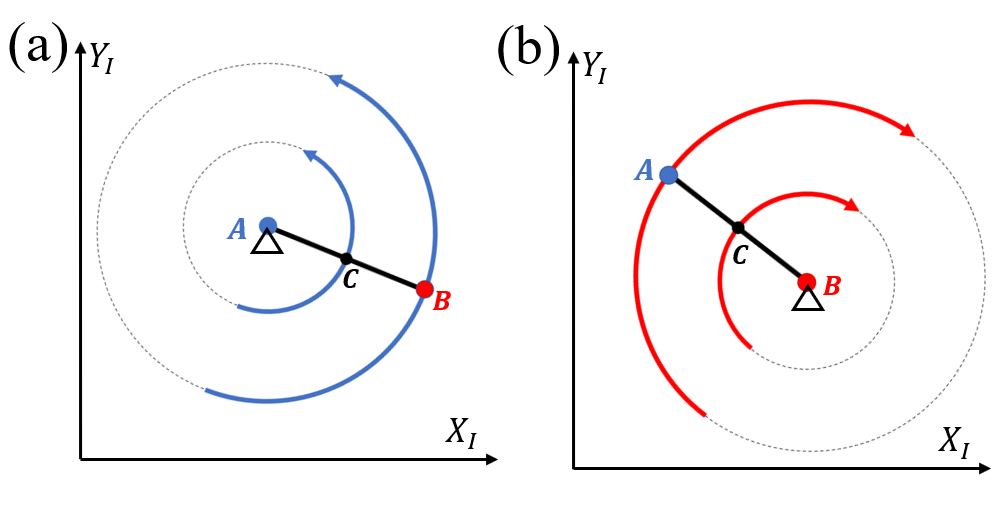}
	\caption{\small{The robot motion about each pivot.}}
	\label{Pivots}
\end{figure}

Thus, we need to obtain the kinematic equations in the millirobot frame. But, before presenting the derivation of the kinematic model of the pivot walker, an assumption should be presented. The robot can only rotate about the pivots without slippage. Figure~\ref{Pivots} describes how the millirobot is rotating about each pivot, wherein Fig.~\ref{Pivots}(a), the active pivot is \textbf{\textit{A}}, therefore this point is fixed and the millirobot is rotating about it. While point \textbf{\textit{B}} is fixed, the pivot \textbf{\textit{B}} is active as shown in Fig.~\ref{Pivots}(b). Successive switching between the two pivot points will enable the millirobot to generate locomotion. 

\normalsize

Since the millirobot has two pivots, it has two kinematic models, one for each pivot. The derivation of the kinematic model for pivot \textbf{\textit{A}} can be presented as follows. The velocity of the millirobot's center in the millirobot frame $\bar{v}_{C}^{R}$ is defined using the relative velocity between points \textbf{\textit{C}} and \textbf{\textit{A}} such as:
\begin{equation}
\bar{v}_{C}^{R} = \bar{v}_{A}^{R} + \bar{\omega} \times \bar{r}_{C/A}^{R}
\label{relA}
\end{equation}
where, $\bar{v}_A$ is equal to zero since the point \textbf{\textit{A}} is fixed pivot, $\bar{\omega}$ is the millirobot angular velocity about $z$-axis, and $\bar{r}_{C/A}^{R}$ is the position vector between point \textbf{\textit{C}} and \textbf{\textit{A}}. Then, $\bar{\omega}$ and $\bar{r}_{C/A}^{R}$ are defined as follows:
\begin{equation}
\bar{\omega} = \begin{bmatrix}
0 \\ 0 \\ \omega
\end{bmatrix}
\end{equation}

\begin{equation}
\bar{r}_{C/A}^{R} = \begin{bmatrix}
0 \\ -\frac{L}{2} \\ 0
\end{bmatrix}
\end{equation}
where, $\omega$ is the angular velocity about the z-axis and $L$ is the millirobot length. So, $\bar{v}_{C}^{R}$ simplifies to 
\begin{equation}
\bar{v}_{C}^{R}  = \begin{bmatrix}
\frac{L \omega}{2} \\ 0 \\ 0
\end{bmatrix}
\end{equation}
Now, using Eq.~(\ref{eq:KM7}), the millirobot velocity in the inertial frame $\bar{v}_{C}^{I}$ can be obtained as follows:
\begin{equation}
\bar{v}_{C}^{R}  = \begin{bmatrix}
\frac{L}{2} \omega \cos (\theta) \\ \frac{L}{2} \omega \sin (\theta) \\ 0
\end{bmatrix}
\label{relA2}
\end{equation}

Similarly, for pivot \textbf{\textit{B}}, Eqs.~(\ref{relA})-(\ref{relA2}) are used to obtain the millirobot velocity in the inertial frame $\bar{v}_{C}^{I}$. Where the relative velocity between the points \textbf{\textit{C}} and \textbf{\textit{B}} is defined as follows
\begin{equation}
\bar{v}_{C}^{R} = \bar{v}_{B}^{R} + \bar{\omega} \times \bar{r}_{C/B}^{R}
\end{equation}
where, $\bar{v}_B$ is equal to zero since the point \textbf{\textit{B}} is fixed pivot and $\bar{r}_{C/B}$ is the position vector between the points \textbf{\textit{C}} and \textbf{\textit{B}}. Then,  $\bar{r}_{C/B}$ is defined as follows:
\begin{equation}
\bar{r}_{C/B}^{R} = \begin{bmatrix}
0 \\ \frac{L}{2} \\0
\end{bmatrix}
\end{equation}

Therefore, the millirobot velocity in the inertial frame $\bar{v}_{C}^{I}$ is described as follows:
\begin{equation}
\bar{v}_{C}^{R}  = \begin{bmatrix}
-\frac{L}{2} \omega \cos (\theta) \\ -\frac{L}{2} \omega \sin (\theta) \\ 0
\end{bmatrix}
\label{relB2}
\end{equation}

The pivot walker kinematic model can be summarized as follows:
\begin{equation}
\dot {\bar{q}}_I = \left\{\begin{matrix}
\left.\begin{matrix}
\dot x =  \frac{L}{2} \omega \cos (\theta)  \\ 
\dot y = \frac{L}{2} \omega \sin (\theta)    \\ 
\dot \theta =  \omega   \\ 
\end{matrix}\right\} &  \text{ for pivot A}\\ 
\\
\left.\begin{matrix}
\dot x =  -\frac{L}{2} \omega \cos (\theta)  \\ 
\dot y = -\frac{L}{2} \omega \sin (\theta)    \\ 
\dot \theta =  \omega   \\ 
\end{matrix}\right\} &  \text{ for pivot B}\\ 
\end{matrix}\right.
\end{equation}
where $\dot x$ and $\dot y$ are the millirobot linear velocities, $\dot \theta$ is the millirobot angular velocity. As can be seen, the kinematic model is a hybrid model that contains continuous-time functions and discrete events (e.g the switching between the two models). Hybrid systems\cite{hurmuzlu2004modeling} are harder to control. As we have control over the switching between the two pivots, the switching can be modeled as input to the kinematic model that results in combining the two kinematic models into a single model as presented:
\begin{equation}
\dot{\bar{q}}_I = \begin{bmatrix}
\frac{L}{2} \sigma \omega \cos (\theta) \\ 
\frac{L}{2} \sigma \omega \sin (\theta)  \\ 
\omega
\end{bmatrix}
\label{KM}
\end{equation}
where $\sigma$ is the switching control input between the models, If  $\sigma = 1$, the pivot \textbf{\textit{A}} is active, whereas  $\sigma = -1$, means that \textbf{\textit{B}} is the active  pivot. Value of the parameter $\sigma$ is constrained by either $1$ or $-1$ depending on the desired active pivot.

Finally, the positions of pivots \textbf{\textit{A}} and \textbf{\textit{B}} are obtained using vector projection. Where pivot \textbf{\textit{A}} position is calculated as follows:
\begin{align} 
x_A &= x - \frac{L}{2} \sin \theta \\
y_A &= y + \frac{L}{2} \cos \theta 
\end{align}
similarly, the position of pivot \textbf{\textit{B}} is
\begin{align} 
x_B &= x + \frac{L}{2} \sin \theta \\
y_B &= y - \frac{L}{2} \cos \theta 
\end{align}
 
\section{Locomotion Analysis} \label{section3}
\label{ControlDesign}
\begin{figure}[!t]
	\centering
	\includegraphics[width=0.75\columnwidth]{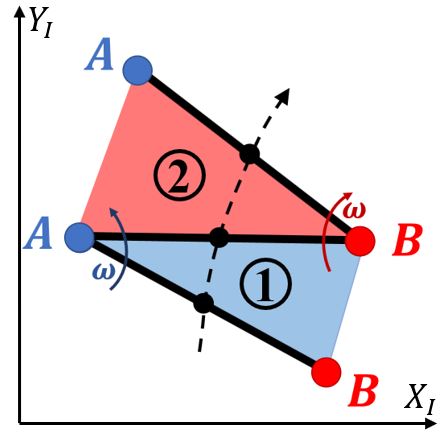}
	\caption{\small{A two-step angular progression in pivot walking. }}
	\label{twoSteps}
\end{figure}

The locomotion of the millirobot is composed of successive steps. Each step can be described as a motion about a pivot point. The location of the pivot point alternates between the two contact points with the ground surface. Figure~\ref{twoSteps} schematically shows a two-step progression of this type of locomotion. During the first step, pivot \textbf{\textit{A}} is active and no motion occurs there. Meanwhile, pivot \textbf{\textit{B}} is free to move with the millirobot body. Once the main link rotates into the desired angle the active pivot is switched to pivot \textbf{\textit{B}} and now pivot \textbf{\textit{A}} is free to move with the main link. This process is repeated to generate forward locomotion.
\normalsize

In general, walking consists of successive steps. To analyze the locomotion of the millirobot, we defined a set of gait parameters. The sweep angle $\theta_s$ is defined as the angle between two successive steps. The number of switches between the two pivots is defined as the number of steps $N_s$. The sweep angle has a significant impact on the motion, the number of steps, and the travel distance. In order to demonstrate the effect of the sweep angle, a locomotion algorithm is designed to control the millirobot to walk along a straight line. This locomotion algorithm was designed such that the millirobot starts from initial position, then takes a step with half of the desired sweep angle. Subsequently, the pivot is switched and followed by a step with a full desired sweep angle. Finally, the pivot is switched again to perform the next step on the other pivot. The straight-line locomotion is shown in Algorithm \ref{StraightWalking}.

\begin{algorithm}[!h]
	\SetAlgoLined
	$N_s =0;$ \, $\theta_0 = 0$\; 
	$\omega = \omega_0;$ \,    $\sigma = 1$ \;
	\While{$y_d-y>0$}{
		\eIf{$N_s = 1$}{
			$\theta_s = \frac{\theta_d}{2} $\;
		}{
			$\theta_s = \theta_d $\;
		}
		\If{$\left | \theta - \theta_0  \right | = \theta_s$}{
			$\sigma = - \sigma $\;
			$\omega= - \omega $\;
			$\theta_0 = \theta$\;
		}
		$q$ = robotKinemaitc ($q$, $\sigma$, $\omega$)\;
		$N_s = N_s +1$\;
	}
	\caption{Straight-line locomotion}
	\label{StraightWalking}
\end{algorithm}

\begin{figure}[!t]
	\centering
	\includegraphics[width=0.96\columnwidth]{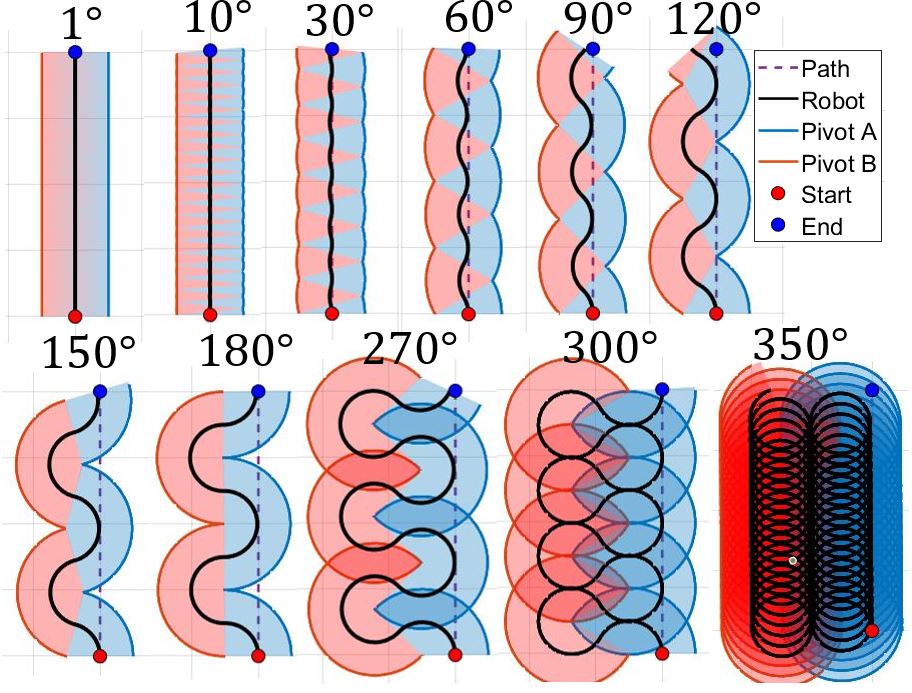}
	\caption{\small{Different robot paths for different sweep angles. The robot length is 10 mm and the desired path length is 40 mm.  }}
	\label{Paths}
\end{figure}

Figure~\ref{Paths} shows different robot trajectories for different sweep angles in the range of $1^{\circ}$ to $360^{\circ}$. Where the purple dashed line represents the straight path, the center of the millirobot trajectory is shown as a thick black line. The blue and red lines represent the pivots \textbf{\textit{A}} and \textbf{\textit{B}} respectively. Blue and red areas indicate that the millirobot rotates about pivots \textbf{\textit{A}} and \textbf{\textit{B}} respectively. The change in color in these areas represents the number of switches between the two pivots, hence the number of steps. As can be seen in Fig.~\ref{Paths}, the sweep angle has a significant effect on the number of steps. Whereas, the number of steps is very large for small angles and gradually reduces for increasing values of the sweep angle. Another observation is that the millirobot path is almost a straight line at smaller angles, while curved paths appear at larger sweep angles. A full parametric analysis is conducted to see the effect of the sweep angle on the number of steps and travel distance. 

\normalsize

\begin{figure}[!b]
	\centering
	\includegraphics[width=1\columnwidth]{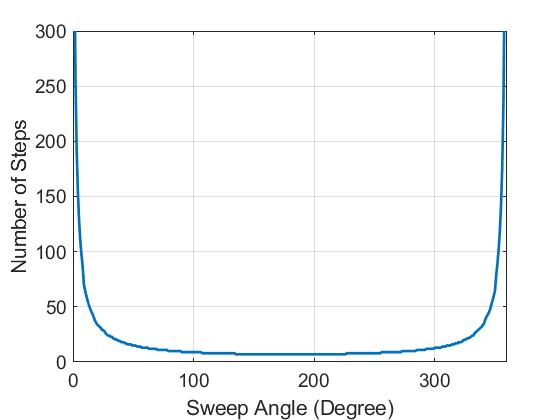}
	\caption{\small{Variation of number of steps to travel 6 cm as a function of sweep angle.}}
	\label{Steps}
\end{figure}	\normalsize

\begin{figure}[!t]
	\centering
	\includegraphics[width=1\columnwidth]{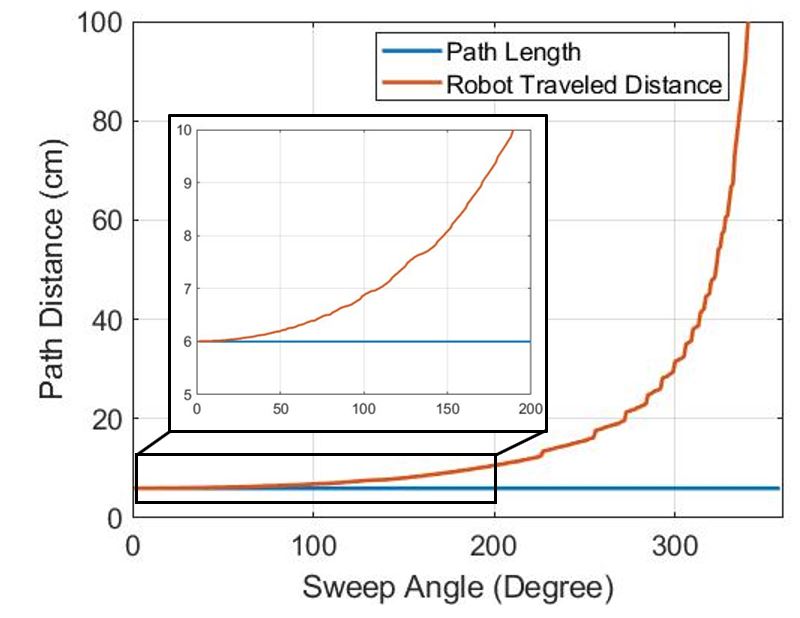}
	\caption{\small{Variation of robot traveled distance as function of sweep angle. }}
	\label{PathDistanceCombined}
\end{figure}

Figure~\ref{Steps} shows the number of steps as a function of the sweep angle for a travel distance of $6$ cm. At lower sweep angles, the number of steps is very large with a singularity at $\theta_s = 0$. Increasing the sweep angle results in a fewer number of steps, As can be seen in Fig.\ref{Steps}, after about $200^\circ$, the number of steps starts increasing as the contribution of the steps to the forward motion diminish. At $\theta_s = 360^\circ$, there is another singularity since the millirobot rotates around its own central axis. Also, a higher number of steps results in longer completion times. Figure~\ref{PathDistanceCombined} depicts the millirobot's travel distance as a function of the sweep angle. Ideally, the travel distance is equal to the path length at lower sweep angles. But increasing the sweep angle results in drifts from the path length leading to higher travel distances, and therefore, to longer task completion times. 

These two parametric analysis results can be used in designing and optimizing control algorithms for the pivot walker robot. For example, a small sweep angle can be used to perform accurate tasks, but small sweep angles result in a higher number of steps. Therefore, the process becomes a trade-off between accuracy and task completion time.

\normalsize

\section{Control Design}\label{section4}
\begin{figure}[!t]
	\centering
	\includegraphics[width=0.75\columnwidth]{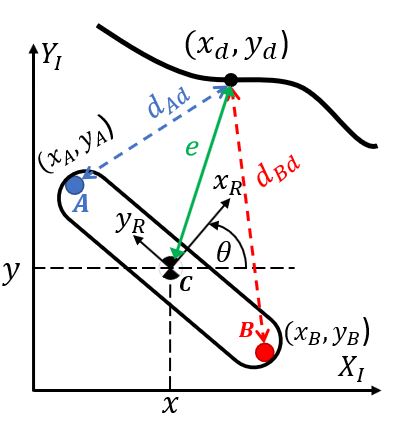}
	\caption{\small{Control problem description.}}
	\label{control1}
\end{figure}

Based on the kinematic model presented in Eq.~(\ref{KM}), the millirobot motion can be governed using two control inputs $\sigma$ and $\omega$. The input $\sigma$ determines the active pivot that the millirobot rotates about and $\omega$ defines how fast the millirobot will rotate about the active pivot. In this section, we present a control algorithm for pivot walking robots to track an arbitrary trajectory. The goal of this controller is to minimize the error $e$ between the millirobot center $(x,y)$ and the desired trajectory $(x_d,y_d)$. Figure~\ref{control1} shows the millirobot and the desired trajectory. The challenge in designing a control algorithm for such systems is how to handle the switching between the two kinematic models. There are two ways to solve this problem. First, pre-plan the motion using algorithms that take into consideration the millirobot kinematics (e.g humanoid and bipedal robots). Second, design an algorithm that can solve the planning and control problem, simultaneously. In this section, we will focus on the second approach, where, two algorithms are proposed to track the desired trajectory without pre-planning the motion.

\normalsize
\subsection{Geometry based controller}
As discussed earlier, the challenge in controlling a walking robot is how to design the control algorithm that switches between the pivots to minimize the error. We propose to use the distance between the two pivots and the desired path, and based on the distance the millirobot uses the nearest pivot as the active pivot and rotates about it. The idea behind this algorithm is that the millirobot rotates about the nearest pivot (e.g. \textbf{\textit{A}}) to the trajectory until the other pivot (e.g. \textbf{\textit{B}}) becomes the nearest one. Then, the pivot is switched to \textbf{\textit{B}} until \textbf{\textit{A}} becomes the nearest pivot to the trajectory, and so on. Consequently, the algorithm is formulated. Expressions for the distances between the pivots and the desired trajectory are shown in Fig.~\ref{control1}, and can be obtained as follows:
\begin{align} 
d_{Ad} &= \sqrt{(x_A-x_d)^2+(y_A-y_d)^2)} \\
d_{Bd} &= \sqrt{(x_B-x_d)^2+(y_B-y_d)^2)}
\end{align}

 Then, the active pivot is:
  \begin{align} 
  \sigma = \text{sign}(d_{Bd}-d_{Ad})
  \end{align}
 where $\sigma = 1$ if $d_{Bd}$ is greater than $d_{Ad}$  and  $\sigma = -1$ if $d_{Bd}$ is less than $d_{Ad}$.
 
 After selecting the active pivot that minimizes the error, a proportional controller for the angular velocity $\omega$ is formulated as follows:
   \begin{align} 
		\omega = k \, \sigma  \, e_n
		\label{omegaLaw}
 \end{align}
 where $k$ is a positive gain, and $e_n$ is the norm of error between the millirobot center and the desired trajectory given by:
   \begin{align} 
	e_n = \sqrt{(x-x_d)^2+(y-y_d)^2)}
	\label{e_n}
 \end{align}
 The sign of $\sigma$ plays an important role in this control law (Eq.~(\ref{omegaLaw})) to specify in what direction the millirobot should rotate. Finally, this control algorithm integrated with the sweep angle concept and shown in Algorithm \ref{controlAlgorithm1}, where the switching between the two pivots is constrained by completing the sweep angle.

\begin{algorithm}[!b]
	\SetAlgoLined
	Initialization \;
	$q=q0$\;
	$e_n = \sqrt{(x-x_d)^2+(y-y_d)^2)}$ \;
	$d_{Ad}= \sqrt{(x_A-x_d)^2+(y_A-y_d)^2)}$ \;
	$d_{Bd}= \sqrt{(x_B-x_d)^2+(y_B-y_d)^2)}$ \;
	$\sigma = \text{sign}(d_{Bd}-d_{Ad})$ \;
	$\omega = \sigma \, e_n$ \;
	\While{$t<t_f$}{

		\If{$\theta - \theta_0 = \theta_s$}{
	$d_{Ad}= \sqrt{(x_A-x_d)^2+(y_A-y_d)^2)}$ \;
	$d_{Bd}= \sqrt{(x_B-x_d)^2+(y_B-y_d)^2)}$ \;
	$\sigma = \text{sign}(d_{Bd}-d_{Ad})$ \;
		}
		$e_n = \sqrt{(x-x_d)^2+(y-y_d)^2)}$ \;
		$\omega = k \, \sigma  \, e_n$ \;
		$q$ = robotKinemaitc ($q$, $\sigma$, $\omega$)\;
		$t =t + \Delta t $\;
	}
	\caption{Geometry based controller}
	\label{controlAlgorithm1}
\end{algorithm}

\subsection{Optimization based controller}
In controls theory, the control law's purpose is to minimize the error between the millirobot and the desired states. In other words, the control problem is an optimization problem with the following cost function $J$. 
\begin{align} 
J =  \frac{1}{2}(q-q_d)^2 
\end{align}
and in more details, 
\begin{align} 
J =  \frac{1}{2}((x-x_d)^2 +(y-y_d)^2+(\theta-\theta_d)^2))
\end{align}
Our goal here is to minimize the cost function, by optimizing the control inputs $\sigma$ and $\omega$. Using the kinematic model (Eq.~(\ref{KM})), the sensitivity of the cost function concerning control inputs is obtained using the chain rule. This control algorithm has a similar structure to a neural network, but it is based on the kinematic network. This network consists of several layers; the first layer is the input nodes and the last layer is the millirobot position and orientation. The intermediate layers represent the kinematic connection between the input and the output. Finally, the back-propagation depends on the same concept of the sensitivity of the objective function for control inputs.

The derivation of the proposed control algorithm is presented next. The sensitivity of the cost function with respect to the angular velocity $\frac{\partial J}{\partial \omega}$ is derived as: 
 \begin{align} 
\frac{\partial J}{\partial \omega} = \frac{\partial J}{\partial x} \frac{\partial x}{\partial \omega} +\frac{\partial J}{\partial x} \frac{\partial x}{\partial \theta} \frac{\partial \theta}{\partial \omega} + \frac{\partial J}{\partial y} \frac{\partial y}{\partial \omega} +\frac{\partial J}{\partial y} \frac{\partial y}{\partial \theta} \frac{\partial \theta}{\partial \omega} +  \frac{\partial J}{\partial \theta} \frac{\partial \theta}{\partial \omega} 
 \end{align} 
 where the partial derivatives are obtained as follows 
 \begin{align} 
 \nonumber
 \frac{\partial J}{\partial x} &= x-x_d  &,
 \frac{\partial J}{\partial y} &= y-y_d  &, 
 \frac{\partial J}{\partial \theta} &= \theta-\theta_d \\ \nonumber
  \frac{\partial x}{\partial \omega} &= \Delta t \, \sigma \cos \theta  &, \frac{\partial y}{\partial \omega} &= \Delta t \, \sigma \sin \theta &, \frac{\partial \theta}{\partial \omega} &= \Delta t \\ 
 \frac{\partial x}{\partial \theta} &= -\Delta t \, \sigma \omega \sin \theta    &, 
 \frac{\partial y}{\partial \theta} &= \Delta t \, \sigma \omega \cos \theta 
  \end{align}
 Next, the sensitivity of the cost function with respect to the active pivot $\frac{\partial J}{\partial \sigma}$ becomes: 
 \begin{align} 
\frac{\partial J}{\partial \sigma} = \frac{\partial J}{\partial x} \frac{\partial x}{\partial \sigma}  + \frac{\partial J}{\partial y} \frac{\partial y}{\partial \sigma} +  \frac{\partial J}{\partial \theta} \frac{\partial \theta}{\partial \sigma}
\end{align} 
  where the partial derivatives are obtained as follows 
 \begin{align} 
 \frac{\partial x}{\partial \sigma} &= \Delta t \, \omega \cos \theta  &, \frac{\partial y}{\partial \sigma} &= \Delta t \, \omega \sin \theta &, \frac{\partial \theta}{\partial \sigma} &= 0 
 \end{align}
Using the gradient decent, the control laws are
  \begin{align} 
 \omega_0 &= \omega_0 + \eta \, \frac{\partial J}{\partial \omega} \\
 \sigma_0 &= \sigma_0 + \eta \, \frac{\partial J}{\partial \sigma}
 \end{align}
 where, $\eta$ is the learning rate. Then, saturation and sign functions are used to constrain the control inputs. 
 \begin{align} 
 \omega &= \text{sat} (\omega_0) \\
 \sigma &= \text{sign} (\sigma_0)
 \end{align}
 
This control algorithm learns and optimizes the control inputs to minimize the cost function, which is the error in this case. Also, this algorithm has an advantage over the previous one because it is able to regulate the orientation of the millirobot. 
 
% Finally, the control algorithm integrated with the sweep angle concept and shown in Algorithm \ref{controlAlgorithm2}. Finally, we note that the switching between the two pivots is constrained by completing the sweep angle.
% 
%\begin{algorithm}[!t]
%	\SetAlgoLined
%	Initialization \;
%	$q=q0$\;
%	$e_n = \sqrt{(x-x_d)^2+(y-y_d)^2)}$ \;
%	$d_{Ad}= \sqrt{(x_A-x_d)^2+(y_A-y_d)^2)}$ \;
%	$d_{Bd}= \sqrt{(x_B-x_d)^2+(y_B-y_d)^2)}$ \;
%	$\sigma = sign(d_{Bd}-d_{Ad})$ \;
%	$\omega = \sigma \, e_n$ \;
%	\While{$t<t_f$}{
%		
%		\If{$\theta - \theta_0 = \theta_s$}{
%			$d_{Ad}= \sqrt{(x_A-x_d)^2+(y_A-y_d)^2)}$ \;
%			$d_{Bd}= \sqrt{(x_B-x_d)^2+(y_B-y_d)^2)}$ \;
%			$\sigma = sign(d_{Bd}-d_{Ad})$ \;
%		}
%		$e_n = \sqrt{(x-x_d)^2+(y-y_d)^2)}$ \;
%		$\omega = k \, \sigma  \, e_n$ \;
%		$q$ = robotKinemaitc ($q$, $\sigma$, $\omega$)\;
%		$t =t + \Delta t $\;
%	}
%	\caption{Geometry based controller}
%	\label{controlAlgorithm2}
%\end{algorithm} 

\section{Simulation Results}\label{section5}

We test the performance of the proposed controllers by conducting a set of simulation runs in different cases. In each case, we use an eight-shape path as the desired trajectory. The initial condition of the millirobot used in the simulation runs is $\bar{q}_I = [0 \text{cm}\,\, -4.2 \text{cm} \,\, \frac{\pi}{2}]^T$, and the effect of varying sweep angle is studied. The $10^\circ$ and $30^\circ$ sweep angles are considered for the first and second cases respectively. In both cases, the following eight-shape trajectory is used.
\begin{align} 
x_d &= -4 \,\sin (0.1 \, t)  \\
y_d &= -4 \,\cos (0.05 \, t)  
\end{align}

It is worth to note that the controllers are tested in the absence of tilt angle and number of steps constraints. Also, the step time is fixed at 0.1 second. 
\begin{figure}[!t]
	\centering
	\includegraphics[width=0.95\columnwidth]{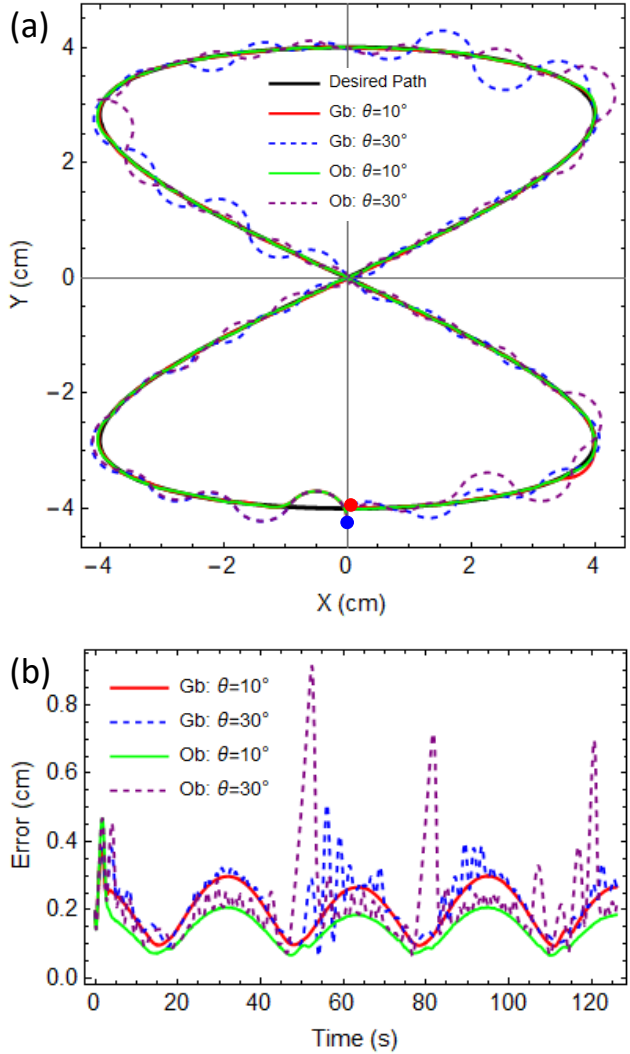}
	\caption{\small{Simulation results of tracking an eight-shape path with different sweep angles. \textbf{(a)} Tracking the desired path with sweep angles 10 and 30 degrees. The Blue and red circles show the start and end positions, respectively. The black, red, dashed blue, green, and dashed purple are the desired trajectory, the simulation results of the geometric based controller of 10 and 30 degrees and the simulation results of optimization based controller of 10 and 30 degrees, respectively. \textbf{(b)} Time histories of the error between the center of the robot and the desired path. In the legend, Gb and Ob denote geometric and optimization based controllers respectively. }}
	\label{Sim}
\end{figure}\normalsize
%\begin{figure}[!b]
%	\centering
%	\includegraphics[width=0.8\columnwidth]{controldeg1_error.JPG}
%	\caption{\small{The norm of the tracking error for a $1^\circ$ sweep angle. The blue and red lines represent the error of method A and B respectively.}}
%	\label{controldeg1_error}
%\end{figure}\normalsize
Figure~\ref{Sim}(a) depicts the simulation results for both cases. As it can be seen, the control methods with smaller $\theta$ perfectly track the desired trajectory. Figure~\ref{Sim}(b) shows the norm of the tracking error $e_n$, which is presented in Eq.~(\ref{e_n}). Therefore, both controllers perform well, but the optimization based controller shows better performance. Also, in terms of the number of steps, the geometric and optimization based controllers are taking 296 and 269 steps to track the desired trajectory, respectively.

%\begin{figure}[!t]
%	\centering
%	\includegraphics[width=1\columnwidth]{controldeg30.JPG}
%	\caption{\small{Pivot walking robot tracking with a $30^\circ$ sweep angle. The green, black and red dashed lines, the red and black circles are the desired trajectory, the millirobot trajectories generated by control method A and B, and the initial and final positions of center of robot, respectively.}}
%	\label{controldeg30}
%\end{figure}
%\normalsize
In order to test the robustness of the proposed controllers, trajectory tracking is tested with a $30^{\circ} $ sweep angle. As one can see in Fig.~\ref{Sim}(a), two control methods track the desired trajectory with more deflections. The norm of the tracking error $e_n$ illustrated in Fig.~\ref{Sim}(b), shows these higher errors. Although two controllers perform well, the optimization based controller outperforms the geometric based one again. Also, in terms of the number of steps, two controllers have approximately a similar number of steps;  $75$ steps are taken in the optimization method and $71$ steps in the geometric one. 

Clearly, the sweep angle affects the controllers' performance because it constrains the millirobot kinematics. As we expected, a comparison of both cases shows that larger sweep angles result in a fewer number of steps but the tracking error performance deteriorates.

%\begin{figure}[!t]
%	\centering
%	\includegraphics[width=0.8\columnwidth]{controldeg30_error.JPG}
%	\caption{\small{The norm of the tracking error for a $30^\circ$ sweep angle.}}
%	\label{controldeg30_error}
%\end{figure}
%\normalsize
\section{Experimental Results}\label{section6}
The experimental results section will be divided into three parts: the first will discuss the experimental setup and signal processing, the second will discuss the tracking performance of the proposed controllers, and the third will discuss the parametric analysis.

\subsection{Experimental setup and signal processing}
\label{ExperimentalSetup}

\begin{figure}[b!]
	\includegraphics[width=1\columnwidth]{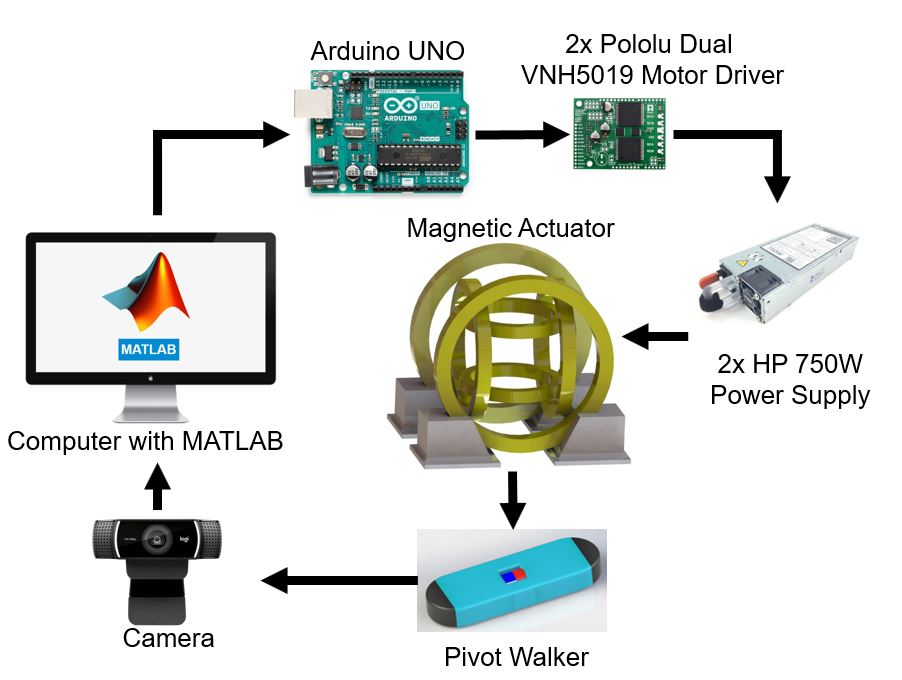}
	\centering
	\caption{Overview of the experimental setup.} 
	\label{OverviewSetup}
\end{figure}
\normalsize

A low-cost high-performance magnetic actuator setup was built to conduct experiments on magnetic actuated milli robots. The setup consists of three nested electromagnetic Helmholtz coils, two HP 750W power supplies, two Pololu Dual VNH5019 motor drivers, Arduino UNO, computer, camera, and pivot walker. The magnetic field was generated using the coils. Figure \ref{OverviewSetup} shows an overview of the experimental setup. The three coils were powered by two HP 750W power supplies which were connected in series. The two coil drivers were used to control the power supplies and transmit the required power to the electromagnetic coils. MATLAB was used to run the main controllers and generate the required input for the coils. Also, MATLAB was used in the feedback to process the image frames from the camera video stream. Arduino board was used as a communication bridge between MATLAB and the coil drivers. Arduino UNO was also used to run the low-level control that converts the high-level commands (e.g $\omega$ and $\sigma$) into current ($I_x$, $I_y$, and $I_z$) command to control the coils.

\begin{figure}[t!]
	\includegraphics[width=1\columnwidth]{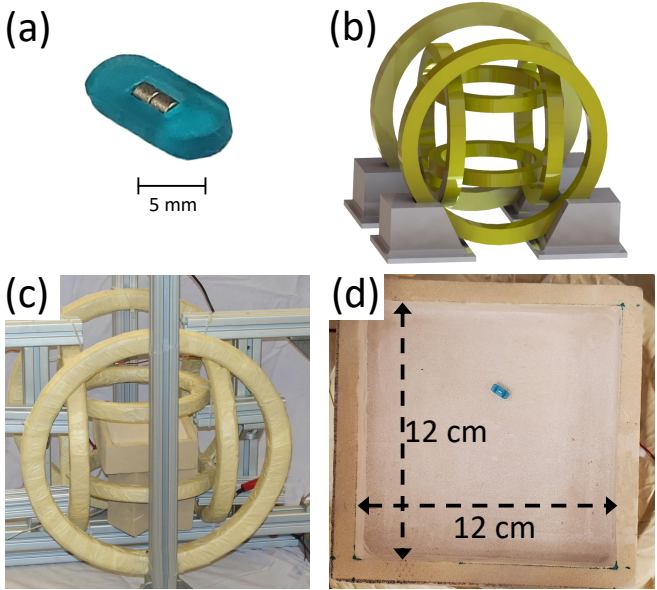}
	\centering
	\caption{\small{ The actual millirobot and experimental set-up.
			\textbf{(a)} 3D-printed 10 mm in length actual millirobot. \textbf{(b)} Isometric view of CAD design of the nested electromagnetic Helmholtz coil. \textbf{(c)} Top view of workspace. \textbf{(d)} Isometric view of experimental set-up. }} 
	\label{coils}
\end{figure}

In the experiments, we use a 10-mm in length millirobot, which is 3D printed using the photo-polymer resin (see Fig.~\ref{coils}(a)). A nested electromagnetic Helmholtz coil actuates this millirobot. The large-scale coil system produces a uniform static magnetic field, which can rotate in a 3D dimension. The outer diameters of coils are 39, 30.5, and 22.5 cm in x, y, and z directions respectively. The separation distances between coil pairs are 24, 19, and 11 cm. Figures~\ref{coils}(b) and (c) show the isometric views of the CAD drawing and the actual coil system. The system has a $12\, \text{cm}\times 12\, \text{cm}$ workspace at the center of the configuration (see Fig.~\ref{coils}(d)). The maximum current applied to the system is eight amps and the system can generate a continuous magnetic field above 10 millitesla (mT).

\normalsize

\begin{figure}[t!]
	\includegraphics[width=1\columnwidth]{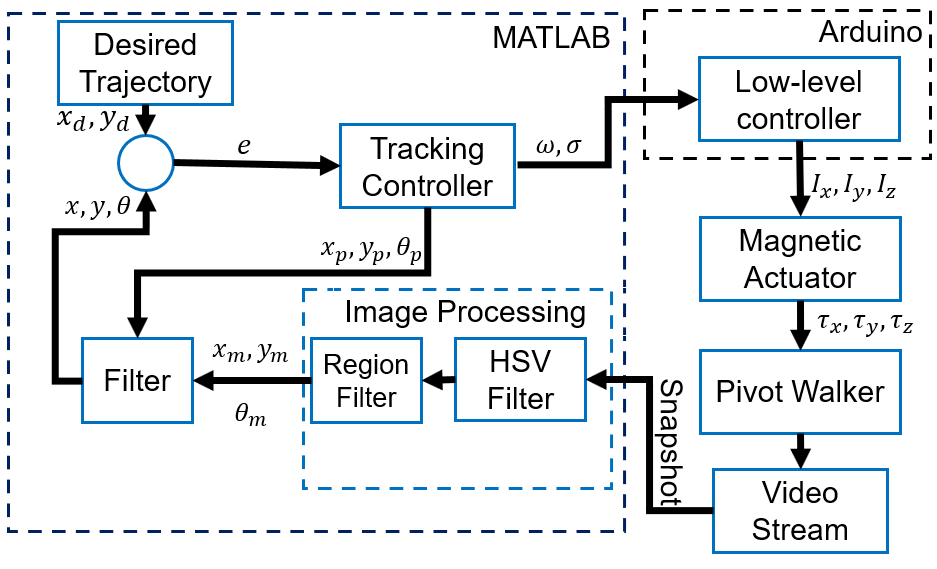}
	\centering
	\caption{The system block diagram.} 
	\label{blockdiagram}
\end{figure}

The pivot walker position and orientation are obtained via visual feedback. Image processing techniques are utilized to estimate the pivot walker pose from the video stream. A snapshot of video steam was taken at every time step. Then, a hue saturation value (HSV) filter was applied to detect the pivot walker color. Subsequently, we use a region filter to detect a rectangle that defines the millirobot based on its area and centroid. Finally,a filter is used to combine the predicted states ($x_p$, $y_p$, and $\theta_p$) from kinematic model with the measured states ($x_m$, $y_m$, and $\theta_m$) from the camera. this filter is helpful to reduce noise and avoid in position and orientation feedback. The overall system structure can be summarized in Fig \ref{blockdiagram}.    

\normalsize

\subsection{Tracking performance}
\label{TrackingPerformance}

In order to validate our proposed control schemes, we conduct a set of different experiments and analyze the data to study the locomotion dynamics. Three different parameters are played the main role in the motion outcomes: tilt angle($\alpha$), sweep angle ($\theta$), and the generating desired path time step ($\Delta t$). We perform parametric analysis on these factors and study the effect of them on the error between the experimental results and the desired path. Here, we define a variable $S$ as a set of the values of three parameters $S=[\theta,\alpha,\Delta t]$; for example if, in an experiment, we use sweep angle as $30^{\circ}$, tilt angle as $15^{\circ}$, and time step as 0.1, the set will be $S=[30,15,0.1]$. Note that the developed kinematic model doesn't take into account the effect of the tilt angle ($\alpha$) so it's quit useful to study the gradual change of this parameter.

To conduct the experimentally parametric analysis on our proposed controllers, we vary the sweep angle between 10 to 40 degrees and the tilt angle between 20 to 30 by increment 5 and choose different values for $\Delta t=\{0.1,0.2,0.3\}$. In total, we have 63 different experimental sets. We consecutively conduct experiments for each set and calculate the summations of the errors between the center of millirobot and the desired trajectory over the entire trajectory. Then, we normalize these summations by the length of the millirobot and compute the mean value denoting by $ME$.  

\begin{figure}[h!]
	\includegraphics[width=1\columnwidth]{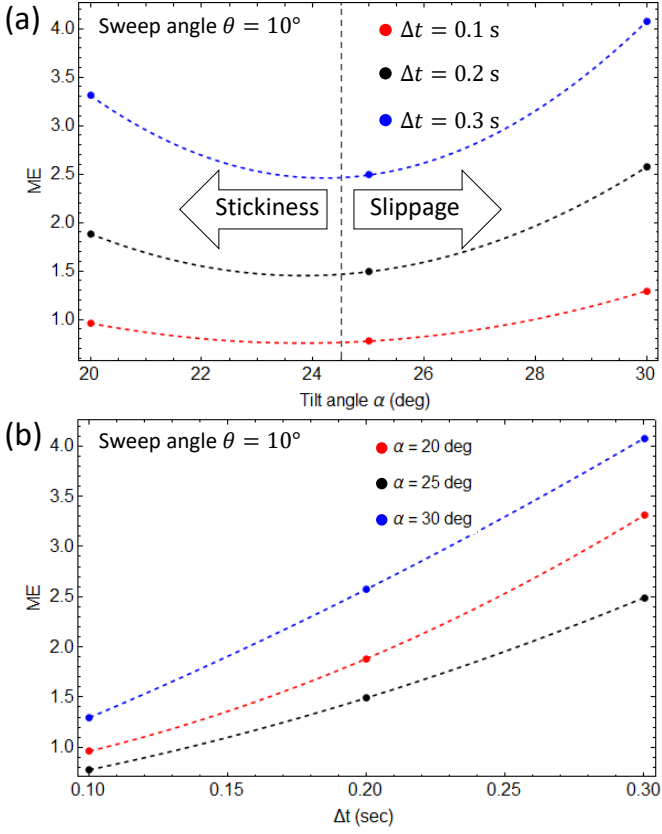}
	\centering
	\caption{\small{ Experimentally parametric analysis of $ME$ when the sweep angle $\theta=10^{\circ}$ is constant.  \textbf{(a)} Variation of $ME$ over different tilt angle. \textbf{(b)} Variation of $ME$ over different step time. }} 
	\label{Para1}
\end{figure}
\normalsize

For the first step, we choose the geometric based controller to analyze the effect of sweep and tilt angles and the time step on $ME$. We conduct ten experiments for each experimental set and obtain the mean values. Figure~\ref{Para1} shows the variation of $ME$ for different tilt angles and step time when the sweep angle is fixed at ten degrees. As one can see in Fig.~\ref{Para1}, the error follows a consistent trend by varying the tilt angle. First, the error is decreasing to reach its minimum value of around $\alpha=24.5^{\circ}$, then it increases. This trend is completely related to the stickiness and the slippage of the pivot points on the surface. 

In Fig.~\ref{Para1}(a), the dashed line in the middle at $\alpha=24.5^{\circ}$ represents the point that the minimum value of the $ME$ occurs. On the left side of this line, the error increases due to more sticky motions on the pivot points. On the other hand, more slippage on the pivot points due to a higher tilt angle causes more error on the right side of the line. Also, for each time step, the slippage problem brings out more error than the stickiness on the pivot point. This pattern can also be seen in Fig.~\ref{Para1}(b). In each time step, the minimum takes place near $\alpha=25^{\circ}$, and tilt angles more than 25 degrees have more error than the angles less than 25 degrees.  

\begin{figure}[h!]
	\includegraphics[width=1\columnwidth]{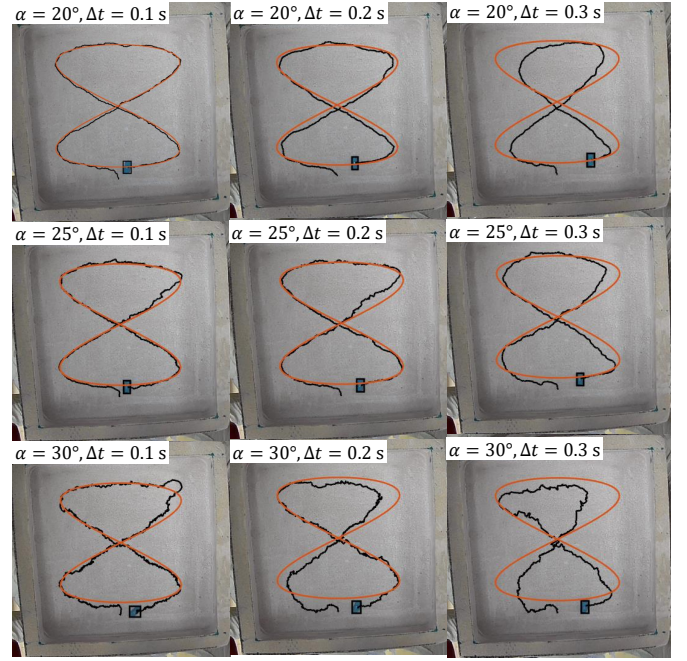}
	\centering
	\caption{\small{Trajectory tracking experimental results when the sweep angle is fixed at $\theta=10^{\circ}$. The orange and black lines are desired trajectory and the actual trajectory of the center of the millirobot respectively. The final position of the millirobot is indicated by a back rectangle at the end of the trajectory.}} 
	\label{expshape1}
\end{figure}
\normalsize 
     
Figure~\ref{expshape1} depicts the actual trajectory tracking of the millirobots, which represents the results of Fig.~\ref{Para1}. In this figure, one can recognize the effects of the stickiness and slippage of the pivot points and varying the time step. As one can see, by increasing the tilt angle, more slippage occurs. By increasing the step time, generating the desired path becomes faster than the reaction of the millirobot, and it can not follow the path. This lag is completely obvious if someone pays attention to the final position of the millirobot. At $\Delta t=0.3 \text{s}$, the distance between the final and initial positions is greater than the others for every tilt angle.      

\begin{figure}[h!]
	\includegraphics[width=0.9\columnwidth]{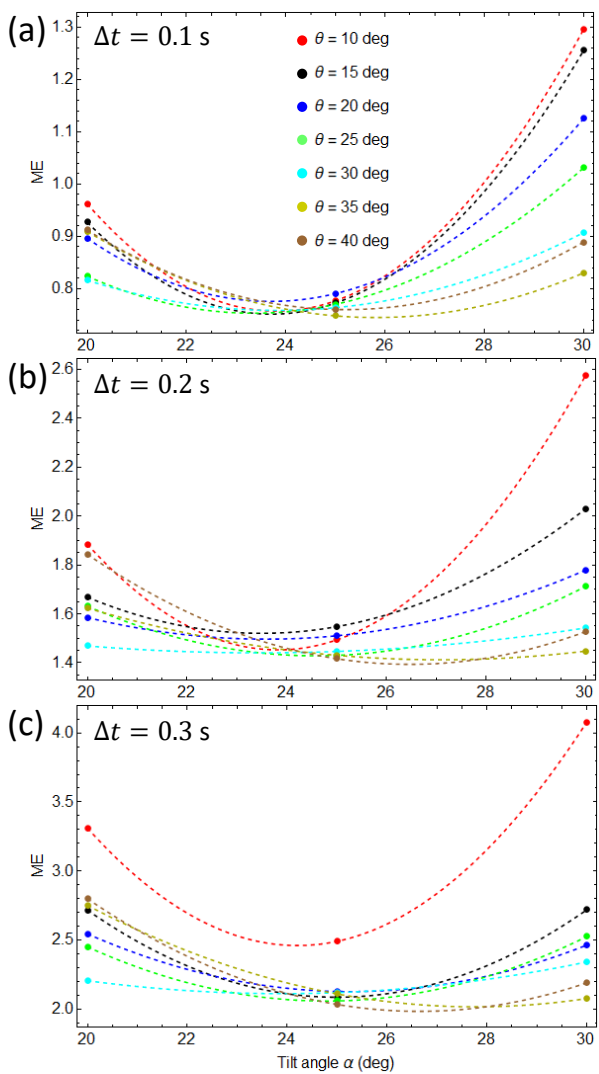}
	\centering
	\caption{\small{ Experimentally parametric analysis of $ME$ when the tilt angle is varying for different sweep angles.\textbf{(a)} Variation of $ME$ for $\Delta t=0.1$ s.\textbf{(b)} Variation of $ME$ for $\Delta t=0.2$ s.\textbf{(c)} Variation of $ME$ for $\Delta t=0.3$ s. }} 
	\label{Para2}
\end{figure}
\normalsize

Figure~\ref{Para2} depicts the parametric analysis over different sweep angles. The minimum $ME$s are near 25 degrees for tilt angle in most of the cases. The surface of the experimental set-up workspace does not have a uniform coefficient of fiction. Thus, we see some inconsistencies in the experimental errors. But, the pattern of increasing in $ME$ happens when the tilt angle is varied up and down. In Fig.~\ref{Para2}, at $\alpha=30^{\circ}$, by increasing the sweep angle, the $ME$ is decreasing. This happens because when the millirobot rotates more in each time step, it can overcome more on the stickiness problem, without adding the slippage problem; but by increasing the time step, the slippage issue appears and as shown in Fig.~\ref{Para2}(a) to (c), the overall mean error increases. The optimum value for sweep angle at $\alpha=30^{\circ}$ is approximately $35^{\circ}$ and at $\alpha=20^{\circ}$ is $30^{\circ}$. We can conclude that the optimum sweep angle can be $30^{\circ}\le \theta_0 \le 35^{\circ} $. 

\begin{figure}[h!]
	\includegraphics[width=0.95\columnwidth]{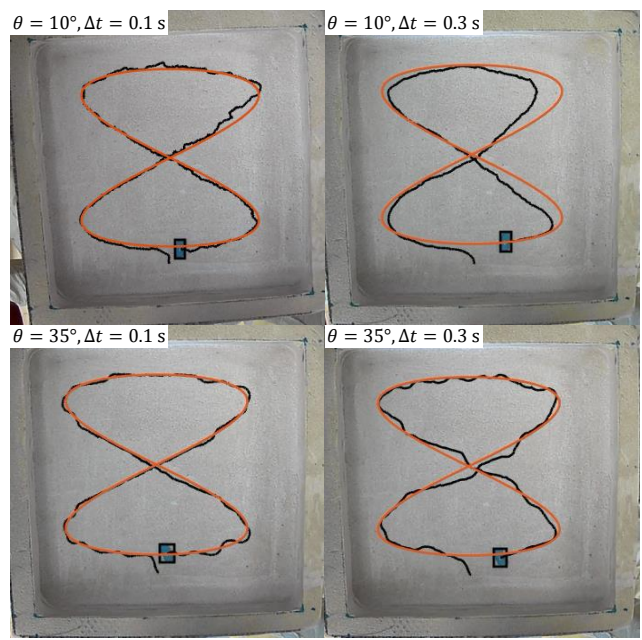}
	\centering
	\caption{\small{Trajectory tracking experimental results for different sweep angles without the tilt angle constraint ($\alpha=25^{\circ}$). The orange and black lines are desired trajectory and the actual trajectory of the center of the millirobot respectively.}} 
	\label{expshape2}
\end{figure}
\normalsize 
 
Figure.~\ref{expshape2} shows the effect of the sweep angle in the trajectory tracking represented in Fig.~\ref{Para2}. One can find the video of these experiments as supplementary materials. 

\begin{figure}[h!]
	\includegraphics[width=1\columnwidth]{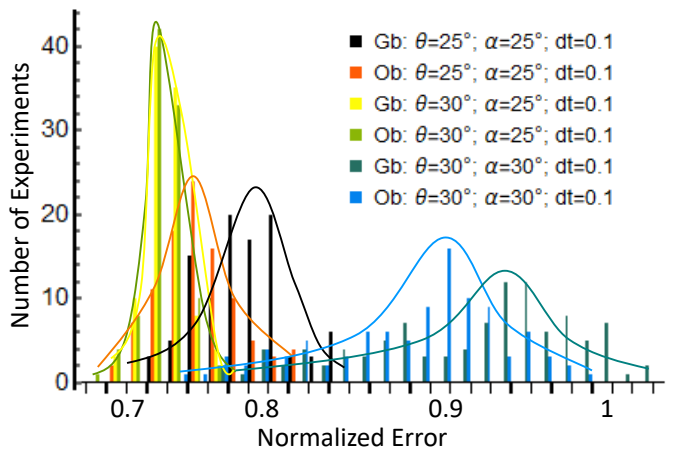}
	\centering
	\caption{\small{The Experimentally error distribution for different controllers. In the legend, Gb and Ob denote geometric and optimization based controllers respectively. }} 
	\label{dis}
\end{figure}
\normalsize

To compare both controllers and analyze the error distribution, we choose three different sets of main parameters and conduct 100 experiments for each set and controller. The sets are: 
\begin{align}
	& S_1=\{25,25,0.1 \} \nonumber\\
	& S_2=\{30,25,0.1 \} \\
	& S_3=\{30,30,0.1 \}\nonumber
\end{align}

\begin{figure}[h!]
	\includegraphics[width=1\columnwidth]{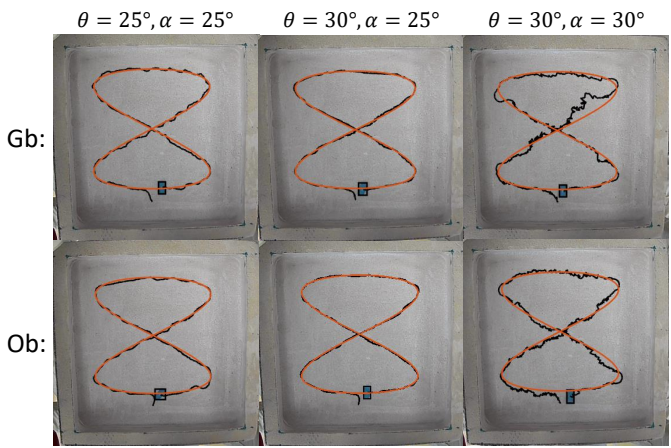}
	\centering
	\caption{\small{Typical trajectory tracking experimental results for different sweep angles and tilt angles, when the time step is constant at $\Delta t=0.1 \text{s}$ and a comparison between two controllers. The first row of pictures belongs to the geometric based controller denoted by Gb. The second row is for optimization based controller (Ob). The experimental set values for each column is placed at the top. The orange and black lines are the desired trajectory and the actual trajectory of the center of the millirobot respectively.}} 
	\label{expshape3}
\end{figure}

We should note that the time step is chosen to be constant in order to check the differentiation of stickiness and slippage factors. Figure~\ref{dis} shows the distribution of normalized error over 100 experiments. As we expect, it shows a semi-bell-shaped curve and more Weibull distribution. The peak of the curve corresponds to the $ME$. As one can see, by increasing the tilt angle, the mean moves to a higher value due to more slippage, and it shows the wider distribution, and smaller peak. Also, a higher sweep angle causes less mean value. Table~\ref{distribution} presents the variance and standard deviation (SD) of each set. In the overall view, as we showed in Fig.~\ref{Sim}(b), the optimal controller has less error than the geometric based controller. It can be concluded via less mean values of bell-curved for the optimal control scheme. Figure.~\ref{expshape3} shows typical examples of these experiments. The out-performance of the optimization based controller can be also recognizable.  

\begin{table}[h!]
	\caption{The variance and standard deviation of experimental results of different sets when the time step is fixed at 0.1 second.} \label{distribution}
	\centering
	\begin{tabular}{|c |c |c |c |c |} \hline 
		Controller& $\theta$ (deg)    & $\alpha$ (deg)  &  Var. &  SD                         \\ \hline 
		Gb & 25 & 25 & 41.8677 & 6.47053 \\ \hline 
		Ob & 25 & 25 & 47.9418 & 6.47625 \\ \hline 
		Gb & 30 & 25 & 98.4696 & 9.92318 \\ \hline 
		Ob & 30 & 25 & 99.291  & 9.96449 \\ \hline 
		Gb & 30 & 30 & 14.0265 & 3.46792 \\ \hline 
		Ob & 30 & 30 & 14.6243 & 3.82418 \\ \hline 
	\end{tabular}
\end{table}

\normalsize  
      
\section{Conclusion}\label{section7}
In this paper, kinematic modeling and control for pivot walking millirobots were addressed. We proposed two control algorithms to guide the pivot walker while it follows the desired path. The first controller was a proportional-geometric based approach. In this approach, the controller activated the nearest pivot point to the desired trajectory. Then, it used a proportional controller to regulate the angular velocity about that pivot point. The second method was based on optimization approach. The gradient descent algorithm was used to optimize the active pivot and the angular velocity and minimize the tracking error between the millirobot and the desired trajectory.

First, we performed a numerical parametric analysis to investigate the effect of the sweep angle on the number of steps and travel distance. We checked the performance of the tracking algorithms within two simulation scenarios, one with 10 degrees sweep angle, the other one with 30 degrees. In both scenarios, two algorithms were tested without placing constraints on the tilt angle and step time.  We showed that the optimization-based controller outperformed the proportional-geometric approach. It had a faster convergence time, smaller tracking error, but a higher number of steps. In addition, we conducted many experimental tests to verify our proposed controllers. We analyzed the effect of sweep angle, tilt angle, and step time in trajectory tracking. We showed that at a certain tilt angle, we have a minimum tracking error. Although in the smaller angle, the stickiness of the pivot point on the ground cause more error, in higher angles, more tracking error is a result of the slippage at the pivot points.

% use section* for acknowledgment
%\section*{Acknowledgment}
%
%
%	This work was partially funded by the National Science Foundation (CMMI 1623324).

% Can use something like this to put references on a page
% by themselves when using endfloat and the captionsoff option.
\ifCLASSOPTIONcaptionsoff
  \newpage
\fi

% trigger a \newpage just before the given reference
% number - used to balance the columns on the last page
% adjust value as needed - may need to be readjusted if
% the document is modified later
%\IEEEtriggeratref{8}
% The "triggered" command can be changed if desired:
%\IEEEtriggercmd{\enlargethispage{-5in}}

% references section

% can use a bibliography generated by BibTeX as a .bbl file
% BibTeX documentation can be easily obtained at:
% http://mirror.ctan.org/biblio/bibtex/contrib/doc/
% The IEEEtran BibTeX style support page is at:
% http://www.michaelshell.org/tex/ieeetran/bibtex/
%\bibliographystyle{IEEEtran}
% argument is your BibTeX string definitions and bibliography database(s)
%\bibliography{IEEEabrv,../bib/paper}
%
% <OR> manually copy in the resultant .bbl file
% set second argument of \begin to the number of references
% (used to reserve space for the reference number labels box)
\bibliographystyle{IEEEtran}
\bibliography{paper}  

% biography section
% 
% If you have an EPS/PDF photo (graphicx package needed) extra braces are
% needed around the contents of the optional argument to biography to prevent
% the LaTeX parser from getting confused when it sees the complicated
% \includegraphics command within an optional argument. (You could create
% your own custom macro containing the \includegraphics command to make things
% simpler here.)
%\begin{IEEEbiography}[{\includegraphics[width=1in,height=1.25in,clip,keepaspectratio]{mshell}}]{Michael Shell}
% or if you just want to reserve a space for a photo:

% You can push biographies down or up by placing
% a \vfill before or after them. The appropriate
% use of \vfill depends on what kind of text is
% on the last page and whether or not the columns
% are being equalized.

%\vfill

% Can be used to pull up biographies so that the bottom of the last one
% is flush with the other column.
%\enlargethispage{-5in}

% that's all folks
\end{document}